\documentclass[sigconf]{acmart}

\AtBeginDocument{%
  \providecommand\BibTeX{{%
    \normalfont B\kern-0.5em{\scshape i\kern-0.25em b}\kern-0.8em\TeX}}}

\setcopyright{acmcopyright}

\copyrightyear{2021}
\acmYear{2021}
\setcopyright{acmcopyright}
\acmConference[KDD '21] {Proceedings of the 27th ACM SIGKDD Conference on Knowledge Discovery and Data Mining}{August 14--18, 2021}{Virtual Event, Singapore.}
\acmBooktitle{Proceedings of the 27th ACM SIGKDD Conference on Knowledge Discovery and Data Mining (KDD '21), August 14--18, 2021, Virtual Event, Singapore}
\acmPrice{15.00}
\acmISBN{978-1-4503-8332-5/21/08}
\acmDOI{10.1145/3447548.3467228}

\usepackage{enumitem}
\usepackage{verbatim}
\usepackage{algorithm}
\usepackage{algorithmic}
\usepackage{mathrsfs}

\usepackage{enumitem}
\usepackage{makecell}
\usepackage{xspace}
 
\usepackage{amsmath,amssymb,amsfonts}
\usepackage{algorithmic}
\usepackage{graphicx}
\usepackage{textcomp}
\usepackage{subfigure}
\usepackage[most]{tcolorbox}
\usepackage{booktabs}
\usepackage{algorithm}
\usepackage{tabularx}
\usepackage{lipsum}
\usepackage{multirow}
\usepackage{color}

\usepackage{CJKutf8}
\usepackage[utf8]{inputenc} 

\usepackage{listings}
\usepackage{xcolor}
\definecolor{light-gray}{gray}{0.99}
\usepackage{caption}
\usepackage{fancybox}
\usepackage[most]{tcolorbox}


\usepackage[T1]{fontenc}
\usepackage{aecompl}

\begin{document}

\title{TabularNet: A Neural Network Architecture for Understanding Semantic Structures of Tabular Data}

\author{Lun Du}
\authornote{These authors contributed equally to the work.}
\authornote{Corresponding Author.}
\affiliation{%
  \institution{Microsoft Research Asia}
  \city{Beijing}
  \country{China}
}
\email{lun.du@microsoft.com}

\author{Fei Gao}
\authornotemark[1]
\authornote{Work performed during the internship at MSRA.}
\affiliation{
  \institution{Beijing Normal University}
  \city{Beijing}
  \country{China}
}
\email{feig@mail.bnu.edu.cn}
 
\author{Xu Chen}
\authornotemark[3]
\affiliation{%
  \institution{Peking University}
  \city{Beijing}
  \country{China}
  }
\email{sylover@pku.edu.cn}

\author{Ran Jia}
\affiliation{%
  \institution{Microsoft Research Asia}
  \city{Beijing}
  \country{China}
}
\email{raji@microsoft.com}

\author{Junshan Wang}
\authornotemark[3]
\affiliation{%
  \institution{Peking University}
  \city{Beijing}
  \country{China}
  }
  \email{wangjunshan@pku.edu.cn}
 
\author{Jiang Zhang}
\affiliation{%
  \institution{Beijing Normal University}
  \city{Beijing}
  \country{China}
 }
\email{zhangjiang@bnu.edu.cn}

\author{Shi Han}
\affiliation{%
  \institution{Microsoft Research Asia}
  \city{Beijing}
  \country{China}
}
\email{shihan@microsoft.com}

\author{Dongmei Zhang}
\affiliation{%
  \institution{Microsoft Research Asia}
  \city{Beijing}
  \country{China}
}
\email{dongmeiz@microsoft.com}

\begin{abstract}
Tabular data are ubiquitous for the widespread applications of tables and hence have attracted the attention of researchers to extract underlying information. One of the critical problems in mining tabular data is how to understand their inherent semantic structures automatically. Existing studies typically adopt Convolutional Neural Network (CNN) to model the \emph{spatial} information of tabular structures yet ignore more diverse \emph{relational} information between cells, such as the hierarchical and paratactic relationships. To simultaneously extract \emph{spatial} and \emph{relational} information from tables, we propose a novel neural network architecture, \textbf{TabularNet}. The spatial encoder of TabularNet utilizes the row/column-level Pooling and the Bidirectional Gated Recurrent Unit (Bi-GRU) to capture statistical information and local positional correlation, respectively. For relational information, we design a new graph construction method based on the WordNet tree and adopt a Graph Convolutional Network (GCN) based encoder that focuses on the hierarchical and paratactic relationships between cells. Our neural network architecture can be a unified neural backbone for different understanding tasks and utilized in a multitask scenario. We conduct extensive experiments on three classification tasks with two real-world spreadsheet data sets, and the results demonstrate the effectiveness of our proposed TabularNet over state-of-the-art baselines.
\end{abstract}


\fancyhead{}

\maketitle
\section{INTRODUCTION}
\begin{figure*}[t]
\includegraphics[width=1.0\textwidth]{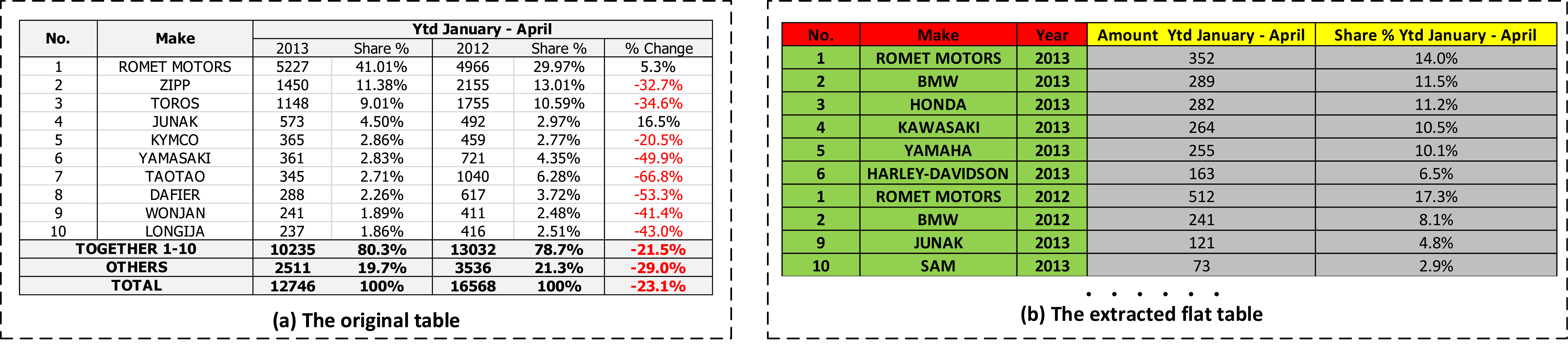}
\caption{An example illustrates transformation of tabular data. (a) is the original table with an intricate structure and (b) shows the targeted formal database table.} 
\label{fig:Intro_case}
\end{figure*}

Tabular data are ubiquitous in spreadsheets and web tables \cite{scaffidi2005estimating,lehmberg2016large} due to the capability of efficient data management and presentation.
A wide range of practical applications for tabular data have been yielded in recent works, such as table search \cite{zhang2019table2vec,lehmberg2015mannheim}, insight chart recommendation \cite{zhoutable2analysis,ding2019quickinsights}, and query answering \cite{tang2015document}. 
Understanding semantic structures of tabular data automatically, as the beginning and critical technique for such applications \cite{dong2019tablesense,dong2019semantic}, aims to transform a ``human-friendly'' table into a formal database table. As shown in Fig. \ref{fig:Intro_case}, the original table (Fig. \ref{fig:Intro_case} (a)) are reformatted to a flat one (Fig. \ref{fig:Intro_case} (b)). Since tables made by human beings play an important role in data presentation, the organizations of tabular data are more complicated for machine understanding. 

\begin{figure*}[t]
\includegraphics[width=1.0\textwidth]{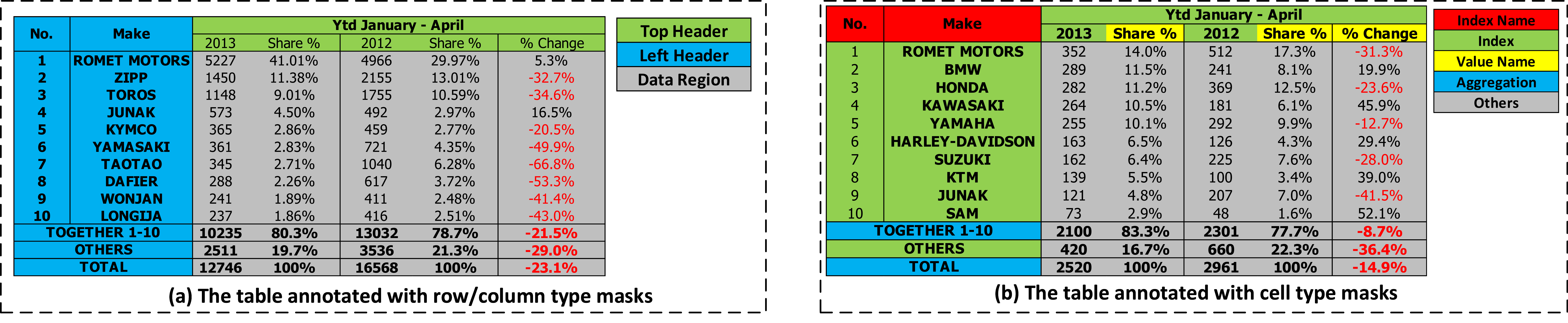}
\caption{An example illustrates two sub-tasks of understanding semantic structure of tabular data. (a) is the region detection task that detects header region and data region of a table. (b) demonstrates the cell classification tasks which classifies each cell into different classes.
} 
\label{fig:Intro_task}
\end{figure*}

The key to such transformation is to understand which cell reflects the name of a data field and which one is a part of a record. The solution of the problem is not so straightforward and is typically divided into some well-defined machine learning sub-tasks \cite{dong2019tablesense}. Two representative sub-tasks, header region detection and cell role classification \cite{dong2019semantic}, are designed to interpret tables from different levels: header region detection, as illustrated in Fig. \ref{fig:Intro_task} (a), detects a group of adjacent cells with header roles; while cell role classification meticulously identifies the semantic role of each cell and it is exemplified in Fig. \ref{fig:Intro_task} (b).

Traditional approaches mainly focus on manually-engineered stylistic, formatting, and typographic features of tabular cells or row/columns \cite{chen2013automatic,koci2016machine}. However, the spatial information, which shows how adjacent cells are organized, is ignored in these methods. Recently, neural network approaches have been very successful in the field of computer vision, natural language processing and social network analysis, and hence have been extensively applied to some specific tasks for tabular data understanding and achieve remarkable results \cite{azunre2019semantic,dong2019tablesense,dong2019semantic,gol2019tabular,nishida2017understanding,zhang2019table2vec,zhang2018smarttable}.
The commonly used neural networks can be roughly divided into \emph{spatial-sensitive} networks and \emph{relational-sensitive} networks. Convolutional Neural Networks (CNN) \cite{he2016deep,szegedy2015going,kalchbrenner2014convolutional} and Recurrent Neural Networks (RNN) \cite{visin2015renet,sutskever2014sequence,liu2016recurrent} are two classic spatial-sensitive networks which aim to capture spatial correlation among pixels or words. Relational-sensitive networks like Graph Convolutional Networks (GCN) \cite{kipf2016semi,xu2018powerful,ijcai2019-606,du2018galaxy} and Self-Attention \cite{vaswani2017attention,devlin2018bert} are designed to extract similarity of node pairs where nodes are highly correlated. Recent studies view tables as image-like cell matrices so that CNNs are widely adopted for the spatial structure correlation \cite{dong2019tablesense,dong2019semantic,azunre2019semantic,nishida2017understanding}, 
leaving the problem of mining relationships between cells unsolved.

For an in-depth understanding of how tabular data are organized and correlated, \emph{spatial} and \emph{relational} information should be considered simultaneously. To be specific, two kinds of \emph{spatial} information should be captured, which include \textbf{statistical information} and \textbf{cross-region variation}. Cells in the same row/column are more semantically consistent from which statistical information such as average, range, and data type can be extracted. On the other hand, there are two different regions in most tables, i.e., a header region and a data region, which is not visible for machines yet extremely vital for the structure understanding. Thus, the spatial-sensitive operation should provide enough cross-region distinguishability. The \emph{relational} information includes hierarchical and paratactic relationships between cells. For instance, cell ``2013'' and ``2012'' in the second row of Fig. \ref{fig:Intro_task} (b) are highly correlated as they play the same role in the header region and both belong to the first-row cell ``Ytd January - April''. A better understanding of such relationships benefits the table transformation process; for example, we know that ``2013'' and ``2012'' should be a part of the records instead of the name of a data field because of their paratactic relationship and the fact that they have a father cell in the hierarchical structure of header region. Obviously, spatial-sensitive networks will fail to capture such complicated \emph{relational} information.

\begin{figure*}[!ht]
\includegraphics[width=1.0\textwidth]{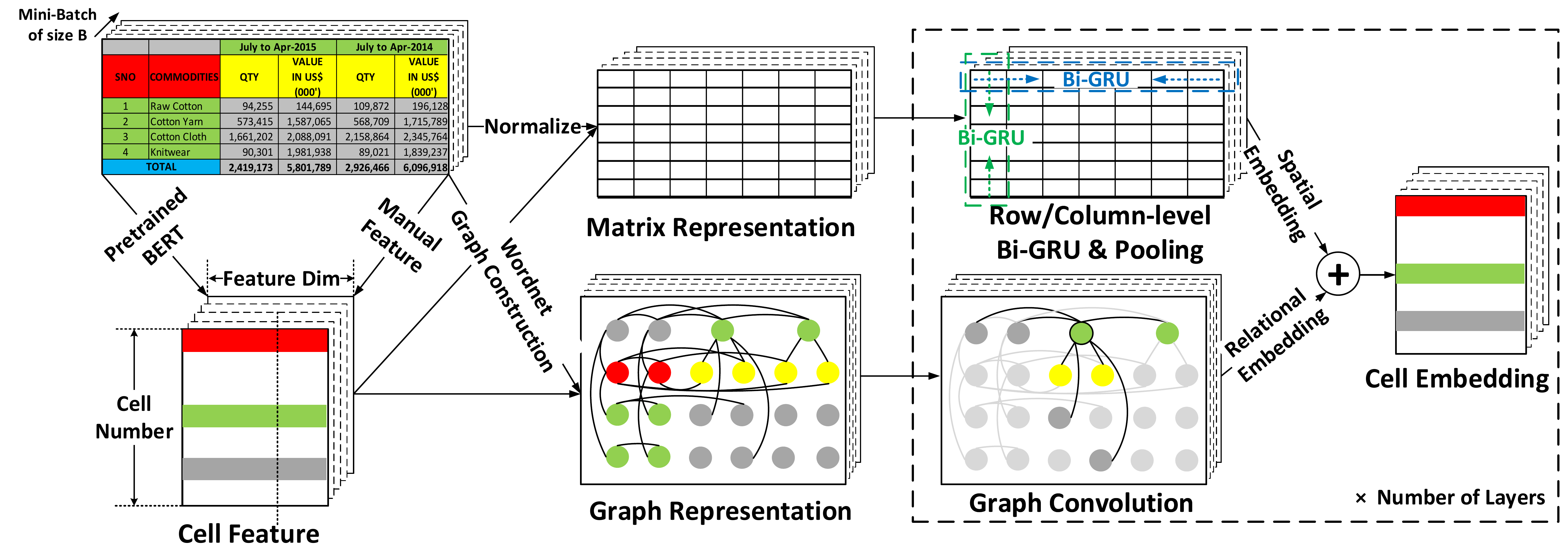}
\caption{An overview of TabularNet. TabularNet is composed of three modules: \textbf{cell-level feature extraction}, \textbf{tabular data representation}, and \textbf{tabular structure information mining}. The cell-level features consist of manual features and semantic embeddings from the pre-trained BERT model. Matrix and Graph representations reveal the spatial information and diverse relationships between cells, respectively. The tabular structure information mining module stacks structure information mining layers, each of which consists of row/column-level Bi-GRU, Pooling, and Graph Convolutional operations.} 
\label{fig:overview}
\end{figure*}

In this paper, we propose a novel neural network architecture \textbf{TabularNet} to capture spatial and relational information of tabular data at the same time. Specifically, tabular data are represented as \emph{matrices} and \emph{graphs} simultaneously, with \emph{spatial-sensitive} and \emph{relational-sensitive} operations well designed for these two representations, respectively. 
 Row/column-level Pooling and Bi-GRU are utilized to extract spatial information from matrix representations for Pooling emphasizes statistics of row/column features and Bi-GRU provides vital distinguishability, especially for cells in region junctions.
 To obtain hierarchical and paratactic relationships of cell pairs from their semantics, we design a WordNet \cite{miller1995wordnet} tree based graph construction algorithm and apply GCN on the graph with cells as nodes and edges connecting related cells.
 Besides, the proposed neural network can be viewed as a unified backbone for various understanding tasks at multiple scales, and hence enable us to utilize it in a multi-task or a transfer learning scenario.
To verify the effectiveness of TabularNet, we conduct two semantic structure understanding tasks, cell role classification and header detection, on two real-world spreadsheet data sets. 
We evaluate TabularNet via comprehensive comparisons against state-of-the-art methods for tabular data based on Transformer, CNN and RNN models, respectively. Extensive experiments demonstrate that TabularNet substantially outperforms all the baselines and its multi-task learning ability further promotes the performance on both tasks. 

The main contributions can be summarized as follows:
\begin{itemize}
 \item We propose a novel neural network architecture TabularNet that can be used as a unified backbone for various understanding tasks on tabular data. 
 \item We introduce the Wordnet tree, GCN and row/column Bi-GRU to simultaneously model relational and spatial information of tabular data.
 \item We conduct header region detection and cell role classification experiments on two real spreadsheet datasets. The results show that TabularNet outperforms the state-of-the-art methods and is adaptive for different tasks as well as effective for the multi-task learning scenario.
\end{itemize}

\section{RELATED WORK}
    There are a considerable number of practical applications for tabular data based on automated semantic structure understanding \cite{zhang2019table2vec,lehmberg2015mannheim,zhoutable2analysis,ding2019quickinsights,tang2015document}. Automated semantic structure understanding methods focus on transforming tabular data with arbitrary data layout into formal database tables. One class of the methods is rule-based, including engineered, \cite{cunha2009spreadsheets,shigarov2016rule}, user-provided \cite{kandel2011wrangler} and automatically inferred \cite{dou2018expandable, abraham2006inferring} techniques. As the amount of available tabular data increases, more machine learning methods are proposed for better generalization performance. The understanding task is split into several well-defined sub-tasks at multiple scales of tables such as table type classification for structure examination \cite{nishida2017understanding}, row/column-level classification for header region detection \cite{dong2019semantic}, and cell-level classification for cell role determination \cite{dong2019tablesense,gol2019tabular}. These methods are composed of two steps: (1) extraction of cell features, including formatting, styling, and syntactic features; (2) modeling table structures. In the first step, most of them use natural language models to extract syntactic features, such as pretrained or fine-tuned RNN \cite{nishida2017understanding} and Transformer \cite{dong2019semantic,dong2019tablesense}. In the second step, spatial-sensitive neural methods are often adopted, such as CNN \cite{dong2019tablesense,dong2019semantic,azunre2019semantic,nishida2017understanding} and RNN \cite{gol2019tabular}. In this paper, we focus on the second step to explore a better neural backbone for different tasks on tabular data.
\section{TabularNet}
In this paper, we propose a neural network architecture \textbf{TabularNet} for table understanding tasks and the overall framework is illustrated in Fig. \ref{fig:overview}. TabularNet is composed of three modules: cell-level feature extraction, tabular data representation, and tabular structure information mining.
The cell-level features consist of handcrafted features and semantic embeddings from BERT, a pre-trained natural language processing model \footnote{The selected features are listed in Appendix.}. 
Matrix and graph representations are both adopted in TabularNet to organize tabular data in spatial and relational formats, respectively. 
Based on these two representations, we design a structure information mining layer that takes advantage of two spatial-sensitive operations, Bi-GRU and Pooling, for matrix representations and a relational-sensitive operation GCN for graph representations. The structure information mining layers are stacked as an encoder to obtain cell embeddings. For various tabular understanding tasks, cell-level embeddings can be purpose-designed at multi-levels. The details of TabularNet will be explained in this section.

\subsection{TabularNet Data Representation}
Traditionally, tabular data are organized as matrices for the application of spatial-sensitive operations as they are analogous with images. However, this representation can not handle the problem of how to reveal the hierarchical and paratactic relationships between cells, which raises the demand for using graphs to represent the tabular data. In this section, We elaborate on the matrix and graph representations of tabular data.
\subsubsection{\textbf{Matrix Representations}}
The great success of CNN on computer vision inspires researchers to apply spatial-sensitive operations on tabular data; thus, tables are organized as matrices to provide spatial information like images. Note that there exist plenty of merged cells due to the need for data formatting and presentation, and it may cause inconvenience when using such operations. A ``Normalize" operation is adopted to split these merged cells into individual cells with the same features and align them so that each matrix element corresponds to a meaningful cell. 

\subsubsection{\textbf{WordNet Based Graph}} 
Although matrix representations shed light on the spatial information of table cells, intricate hierarchical and paratactic correlations in the header region are still less considered in recent studies. Cells which are peer groups play similar roles and can be easily distinguished by human beings from their texts. Nonetheless, rapidly varying cell roles with positions in header region raise the problem of extracting such relationships from matrix representations for these cells may be far apart. To address this difficulty, we introduce WordNet \cite{miller1995wordnet} to construct graph representations under the semantics of cell texts. WordNet is a lexical database of semantic relations between words where synonyms are linked to each other and eventually grouped into synsets. Nouns in WordNet are organized into a hierarchical tree where words at the same level share similar properties. Our empirical study on spreadsheet data set of \cite{dong2019semantic} show that $50.2\%$ header cells have texts and only $6.5\%$ data cells have texts. Based on the intuition that cells that are peer-group correlated would present similarity in cell texts, we proposed a WordNet based graph representations for tables:
 \begin{algorithm}
 \caption{Word Set}\label{alg:word set}
\begin{algorithmic}
\REQUIRE{Table $X$, Cell $x$, synset $S(w)$ of word $w$ from WordNet, the maximum number of synonyms $\eta$}
\ENSURE Word set $WS(x)$.
\FOR{Cell $x$ in tabel $X$}
 \FOR{Word $w_i$ in text of cell $x$}
 \STATE{$WS(x)$.append($S(w_i)[0],\cdots,S(w_i)[\eta-1]$)}
 \ENDFOR
\ENDFOR
\end{algorithmic}
\end{algorithm}
 
 \begin{algorithm} 
 \caption{WordNet Based Graph Construction}\label{alg:graph}
 \begin{algorithmic}[1]
 \REQUIRE{Table $X$ , WordNet Tree $T$, depth gap threshold $\epsilon$}
 \ENSURE Graph $G$.
 \STATE{Word Set(X)}
 \FOR{word $i$ in $WS(x)$ of cell $x$}
 \FOR{word $j$ in $WS(y)$ of cell $y$}
 \IF{word pair $(i,j)$ in the same layer of WordNet Tree $T$ \textbf{and} the lowest common ancestor of $(i,j)$ is within $\epsilon$ layers higher than $(i,j)$}
 \STATE{$G$.add\_edges($i,j$)}
 \ENDIF
 \ENDFOR
 \ENDFOR
 \RETURN $G$
 \end{algorithmic}
\end{algorithm}
Algorithm. \ref{alg:graph} demonstrates that word pairs from different cells are considered highly correlated if they are in the same layer of WordNet tree $T$. To avoid ambiguity while reducing the consuming time in graph construction, we utilize the synsets of each word with the limitation that the number of synonyms is less than $\eta$. Note that all words have a common ancestor ``Entity" as it is the root of $T$, we apply a constraint that the gap between the depth of word pairs and their lowest common ancestor should less than a threshold $\eta$ to avoid too diverse semantics of word pairs. 

\subsection{Structure Information Mining Layer}
\textbf{Spatial Information Encoder.} 
There is abundant underlying spatial information in matrix representations of tables, requiring proper design of spatial-sensitive operations. Motivated by the fact that cells in the regional junction are significant for the detection of header region and data region, we firstly utilize a row-level and a column-level Bi-GRU \cite{ChoMGBBSB14} with different parameters to capture spatial dependencies of cells in the same rows and the same columns, respectively. A Bi-GRU observes a sequence of input vectors of cells and generates two hidden state sequences as outputs. Each hidden state sequence extracts one direction spatial information of the input sequence. Since the header region is usually at the top or the left side of a table, the information captured by row-level Bi-GRU starting from the left and the right will show great differences in cells in the regional junction, and so it is with column-level Bi-GRU. It indicates that Bi-GRU is able to provide substantial distinguishability, especially for cells in the regional junction.

Denote the feature for cell $\mathcal{C}_{ij}$ at position $(i, j)$ as $\mathbf{x}_{ij}$, and the cell sequence in each row becomes the input of row-level Bi-GRU. The row-level Bi-GRU will generate two hidden state vectors $\mathbf{h}^l_{ij}$ and $\mathbf{h}^r_{ij}$ that memorize important spatial information from the left and the right side to cell $\mathcal{C}_{ij}$, respectively.

Similarly, through the column-level Bi-GRU, we can obtain another two hidden state vectors $\mathbf{h}^t_{ij}$ and $\mathbf{h}^b_{ij}$ capturing spatial dependencies in the column direction starting from the top and bottom direction separately. The final Bi-GRU representations $\mathbf{h}^g_{ij}$ can be formulated as:
\begin{align}
 \mathbf{h}^g_{ij} = \sigma \left(\mathbf{W}^b_r (\mathbf{h}^l_{ij} \oplus \mathbf{h}^r_{ij}) + \mathbf{b}^b_r \right) \oplus \nonumber 
 \sigma\left(\mathbf{W}^b_c (\mathbf{h}^b_{ij} \oplus \mathbf{h}^t_{ij}) + \mathbf{b}^b_c \right),
\end{align}%
where $\mathbf{W}^b_r$, $\mathbf{W}^b_c$, $\mathbf{b}^b_r$, and $\mathbf{b}^b_c$ are learnable parameters for the two Bi-GRU operation, ``$\oplus$'' is the concatenation operation and $\sigma$ is the ReLU activation function.

Apart from Bi-GRU, we adopt row/column-level average Pooling to get the statistics of each row/column because the informative statistics reveal the properties of most cells in a row/column. It is especially beneficial when classifying cells in the header region.
The row and column level representations from Pooling operations can be formulated as follows:
\begin{equation}
 \begin{split}
 \mathbf{r}_i = \frac{1}{N_c}\sum_{j=1}^{N_c} \mathbf{MLP}(\mathbf{x}_{ij}), \quad
 \mathbf{c}_j = \frac{1}{N_r}\sum_{i=1}^{N_r} \mathbf{MLP}(\mathbf{x}_{ij}),
 \end{split}
\end{equation}
where $\mathbf{r}_i$ is embedding of the $i$-th row and $\mathbf{c}_j$ is embedding of the $j$-th column. $N_r$ represents the length of each row and $N_c$ denotes the length of each column. $\mathbf{MLP}$ is a multilayer perceptron with ReLU as activation fucntion.

Eventually, we concatenate the row, column and cell embeddings together as the spatial embedding for cell $\mathcal{C}_{ij}$:
\begin{equation}
 \mathbf{h}_{ij}^s = \mathbf{h}_{ij}^g \oplus \mathbf{r}_i \oplus \mathbf{c}_j.
\end{equation}

\textbf{Relational Information Encoder.} Based on the constructed graph, we utilize a Graph Convolutional Network (GCN) \cite{kipf2016semi} to capture the diverse relationships between cells. GCN is a extensively used neural network to encode relationships of node pairs on graph structure data and have achieved significant improvement on corresponding tasks. Specifically, we modify Graph Isomorphism Network (GIN) \cite{xu2018powerful}, a powerful variant of GCN, to capture the relationships. GIN aggregates information from neighbor nodes and passes it to the central node, which can be represented as:
\begin{equation}
    \label{equ:GIN}
    \resizebox{.91\linewidth}{!}{$
    \displaystyle
    \mathbf{h}_{ij}^r(l+1) = \mathbf{MLP}^o \left ((1+\epsilon) \cdot \mathbf{h}_{ij}^r(l) + \sum_{\mathcal{C}_{kt} \in \mathcal{N}(\mathcal{C}_{ij})} \mathbf{h}_{kt}^r(l)\right),
  $}
\end{equation}

\begin{equation}
    \mathbf{h}_{ij}^r(0) = \mathbf{MLP}^e(\mathbf{x}_{ij}),
\end{equation}
where $\mathcal{N}(\mathcal{C}_{ij})$ is neighbor node set of the central cell $\mathcal{C}_{ij}$ on the WordNet based graph, and $\mathbf{\epsilon}$ is a learnable parameter. $\mathbf{h}_{ij}^r(l)$ is the relational embedding of the cell $\mathcal{C}_{ij}$ at the $l$-th layer of GIN. $\mathbf{MLP}^e$ and $\mathbf{MLP}^o$ are multilayer perceptrons with different parameters.

In the graph construction algorithm, cells in the data region may have less edges connecting to them or even become isolated nodes. Considering an extreme case that there are all isolated nodes on the graph, the learned relational embeddings for $\mathcal{C}_{ij}$ from Eq. \eqref{equ:GIN} can be rewritten as:
\begin{equation}
 \mathbf{h}_{ij}^r(l+1) = \mathbf{MLP}^o \left ((1+\epsilon) \cdot \mathbf{h}_{ij}^r(l) \right),
\end{equation}
which is approximately equivalent to a multilayer perceptrons. It ensures that the embedding for each cell is in the same vector space. 

The spatial and relational information encoder are combined as a structure information mining layer, which can be stacked for better performance. 

\subsection{Embedding Integration for Various Tasks}
The various tabular understanding tasks require us to learn representations of tabular data at multi-levels. 
Based on the output embeddings from stacked structure information mining layers, we use some simple integration methods to obtain the final representations.

For cell-wise tasks, we directly concatenate the outputs of the last spatial information encoder and relational information encoder as the final representation $\mathbf{h}_{ij}$ of cell $\mathbf{\mathcal{C}}_{ij}$:
\begin{equation}
 \mathbf{h}_{ij} = \mathbf{h}_{ij}^s \oplus \mathbf{h}_{ij}^r.
\end{equation}

Row/column representations are more practical for tasks like region detection. With the cell representation $h_{ij}$, we utilize a mean pooling operation over cells in a row $\mathbf{Row_{i}}$ or column $\mathbf{Col_{k}}$ to obtain the row/column representation $\mathbf{h}_{i}^{row}$ or $\mathbf{h}_{j}^{col}$:
\begin{equation}
\mathbf{h}_{i}^{row} = \frac{1}{N_c} \sum_{j=1}^{N_c} \mathbf{h}_{ij}, \quad \mathbf{h}_{j}^{col} = \frac{1}{N_r}\sum_{i=1}^{N_r} \mathbf{h}_{ij}.
\end{equation}
We use negative log-likelihood as the loss function for both tasks.

For multi-task learning, the cell embeddings or the row/column embeddings can be fed to different decoders with different task-specific losses, and the losses are added directly as the final loss. 

\begin{table*}[h]
\centering
\setlength{\abovecaptionskip}{1pt}%
\setlength{\belowcaptionskip}{5pt}%
\caption{F1-score for cell classification}
\label{cell-class-res}
\begin{tabular}{l c c c c c c c}
\toprule
& Macro-F1 score & Index Name  & Index & Value Name & Aggregation\\
\toprule

SVM & 0.678 $\pm$ 0.008         & 0.557 $\pm$ 0.001 & 0.721 $\pm$ 0.007 & 0.649 $\pm$ 0.010 & 0.483 $\pm$ 0.09\\
CART & 0.612 $\pm$ 0.004 & 0.502 $\pm$ 0.002 & 0.703 $\pm$ 0.001 & 0.562 $\pm$ 0.002 & 0.319 $\pm$ 0.003\\
\hline
FCNN-MT (Our features)  & 0.714 $\pm$ 0.003 & 0.599 $\pm$ 0.001 & 0.777 $\pm$ 0.001 & 0.681 $\pm$ 0.001 & 0.538 $\pm$ 0.001 \\
RNN-PE (Our features)  & 0.685 $\pm$ 0.004 & 0.554 $\pm$ 0.047 & 0.744 $\pm$ 0.009 & 0.605 $\pm$ 0.036 & 0.340 $\pm$ 0.025 \\
RNN-PE & 0.703 $\pm$ 0.005   & 0.516 $\pm$ 0.040 & 0.789 $\pm$ 0.006 & 0.638 $\pm$ 0.027 & 0.596 $\pm$ 0.023 \\
TAPAS (Our features)   & 0.747 $\pm$ 0.007  & 0.672 $\pm$ 0.023 & 0.812 $\pm$ 0.005 & 0.681 $\pm$ 0.006 & 0.589 $\pm$ 0.008 \\
\hline
TabularNet (w/o Bi-GRU)  & 0.732 $\pm$ 0.002 & 0.623 $\pm$ 0.006 & 0.813 $\pm$ 0.005 & 0.684 $\pm$ 0.005 & 0.587 $\pm$ 0.003\\
TabularNet (w/o GCN) & 0.776 $\pm$ 0.006 & 0.701 $\pm$ 0.010 & 0.835 $\pm$ 0.019 & 0.715 $\pm$ 0.008 & 0.639 $\pm$ 0.013\\
TabularNet       & 0.785 $\pm$ 0.010 & 0.704 $\pm$ 0.018 & 0.839 $\pm$ 0.020 & 0.737 $\pm$ 0.014 & 0.671 $\pm$ 0.008\\
TabularNet (Multi-task) & \textbf{0.788 $\pm$ 0.010} & \textbf{0.709 $\pm$ 0.018} & \textbf{0.842 $\pm$ 0.020} & \textbf{0.741 $\pm$ 0.012} & \textbf{0.678 $\pm$ 0.012}\\ 
\bottomrule
\end{tabular}
\end{table*} 

\section{Experiments}
\subsection{Dataset}

We employ two real-world spreadsheet datasets from \cite{dong2019semantic} for two tasks, respectively. Large tables with more than 5k cells are filtered for the convenience of mini-batch training. The dataset for cell classification has $3,410$ tables with $814$k cells and five cell roles (including \textit{Index}, \textit{Index Name}, \textit{Value Name}, \textit{Aggregation} and \textit{Others}). Definitions of five cell roles are described as follows: 

\begin{itemize}[leftmargin=10pt]

\item \textit{Index name}: the name to describe an index set is called an index name. E.g., ``country'' is the index name of (China, US, Australia).

\item \textit{Index}: we define the index set as a set of cells for indexing values in the data region. For example, ``China, US and Australia'' constitute an index set, ``1992, 1993, 1994, 1995, …, 2019'' is also an index set. The elements in the index set are called indexes.

\item \textit{Value name}: a value name is an indicator to describe values in the data region. A value name can be a measure such as ``number'', ``percent'' and ``amount'', and can also be a unit of measure, such as ``meter'' and ``mL''.

\item \textit{Value}: values lie in the data region.

\item \textit{Aggregation name}: an aggregation name indicates some values are calculated from other values. A ``total'' is a special aggregation indicating the sum operation.
\end{itemize}
In \cite{dong2019semantic}, 3 classes (\textit{Index}, \textit{Index Name} and \textit{Value Name}) are selected for the cell classification task, however, all the 5 classes are used in this work for a comprehensive comparison. The dataset for table region detection has $6,988$ tables including $2.9$M cells in total.

Both datasets face the problem of imbalanced class distribution.
In the former dataset, the cell number of type \textit{Others} are nearly more than 300 times that of type \textit{Index Name} or \textit{Aggregation}. Nearly $87.2\%$ of cells in the latter dataset are labeled as \textit{Others}, and only $3.7\%$ of rows and $13.3\%$ of columns are labeled as \textit{Header}.

\begin{table}[h]
\centering
\setlength{\abovecaptionskip}{1pt}%
\setlength{\belowcaptionskip}{5pt}%
\caption{F1-score for table region detection}
\label{region-class-res}
\scalebox{1.0}{
\begin{tabular}{l c c}
\toprule
             & Top Header & Left Header  \\ \hline
SVM             & 0.850 $\pm$ 0.003   & 0.860 $\pm$ 0.010\\
CART             & 0.874 $\pm$ 0.003   & 0.881 $\pm$ 0.003 \\\hline
FCNN-MT (Our Features)    & 0.894 $\pm$ 0.003  & 0.887 $\pm$ 0.003 \\
TAPAS (Our Features)     & 0.909 $\pm$ 0.015  & 0.871 $\pm$ 0.008\\\hline
TabularNet(w/o Bi-GRU)     & 0.896 $\pm$ 0.018  & 0.879 $\pm$ 0.015\\
TabularNet(w/o GCN)    & 0.928 $\pm$ 0.006 & 0.906 $\pm$ 0.024\\
TabularNet          & 0.933 $\pm$ 0.023 & 0.912 $\pm$ 0.009\\
TabularNet(Multi-task)    & \textbf{0.940 $\pm$ 0.014}   & \textbf{0.921 $\pm$ 0.013} \\
\toprule
\end{tabular}
}
\end{table}

\subsection{Task} 
We conduct our experiments on two tasks: \textit{cell role classification} and \textit{region detection}. The first task can be regarded as a cell-level multi-classification problem. The region detection task aims to determine whether a row belongs to the top-header region, and a column belongs to the left-header region, which can be formulated as row-level and column-level binary classification problems.

\subsection{Baselines} 
To verify the effectiveness of TabularNet, we choose five methods as the baselines. 
\begin{itemize}
\item \textbf{SVM} \cite{wenthundersvm18} and \textbf{CART} \cite{cart} are two widely used classifiers before deep learning era.
\item \textbf{RNN-PE} \cite{gol2019tabular} and \textbf{FCNN-MT} \cite{dong2019semantic} are two state-of-the-art NN-based methods for table understanding tasks. RNN-PE is an RNN-based method for cell classification while FCNN-MT is a CNN-based method for both cell classification and region detection. 
\item \textbf{TAPAS} \cite{herzig2020tapas} is a pretrained transformer-based method for QA task on tabular data. We leverage its table representation part as the table encoder with our extracted features and the task-specific loss function for the comparison. Because it is necessary to explore the representation ability of transformer-based methods for a comprehensive comparison, whereas there is no such method for similar tasks. 
\end{itemize}

\subsection{Settings}
Two datasets are randomly split into training, validation and test set with a ratio of 7:1:2.
The models for cell classification and region detection share the same TableNet architecture as their backbone, which consists of one structure information mining layer. Three fully connected layers are adopted as the decoder with cross-entropy loss. The dimensions of embedding vectors in GCN and Bi-GRU are both 128. We set the batch size to 10 (tables) and a maximum of 50 epochs for each model training. 
AdamW \cite{loshchilov2017decoupled} is chosen as the optimizer with a learning rate of $5e-4$ for a fast and stable convergence when training TabularNet.
To prevent over-fitting, we use Dropout with drop probability $0.3$, weight decay with decay ratio $5e-5$, and early stopping with patient $8$. 
For baselines, we follow the default hyper-parameter settings in their released code. The results shown in the following tables are averaged over 10 runs with a fixed splitting seed. The experiments are compiled and tested on a Linux cluster (CPU:Intel(R) Xeon(R) CPU E5-2690 v4 @ 2.60GHz, GPU:NVIDIA Tesla V100, Memory: 440G, Operation system: Ubuntu 16.04.6 LTS). TabularNet is implemented using Pytorch and Pytorch-Geometric.

\begin{table*}[h]
\centering
\caption{Results of cell classification based on different graph construction method.} \label{tableedge}
\begin{tabular}{lcccccc}
\toprule

Graph   & Index Name  & Index & Value Name & Aggregation \\\hline
Grid  & 0.667 $\pm$ 0.012 & 0.823 $\pm$ 0.014 & 0.672 $\pm$ 0.022 & 0.641 $\pm$ 0.023 \\
TlBr  & 0.689 $\pm$ 0.011 & 0.829 $\pm$ 0.023 & 0.697 $\pm$ 0.011 & 0.651 $\pm$ 0.019 \\
Wordnet & \textbf{0.704 $\pm$ 0.018} & \textbf{0.839 $\pm$ 0.020} & \textbf{0.737 $\pm$ 0.014} & \textbf{0.671 $\pm$ 0.008}\\ 
\bottomrule
\end{tabular}
\end{table*}

\begin{figure*}
\centering
\scalebox{0.95}{
\includegraphics[width=1\textwidth]{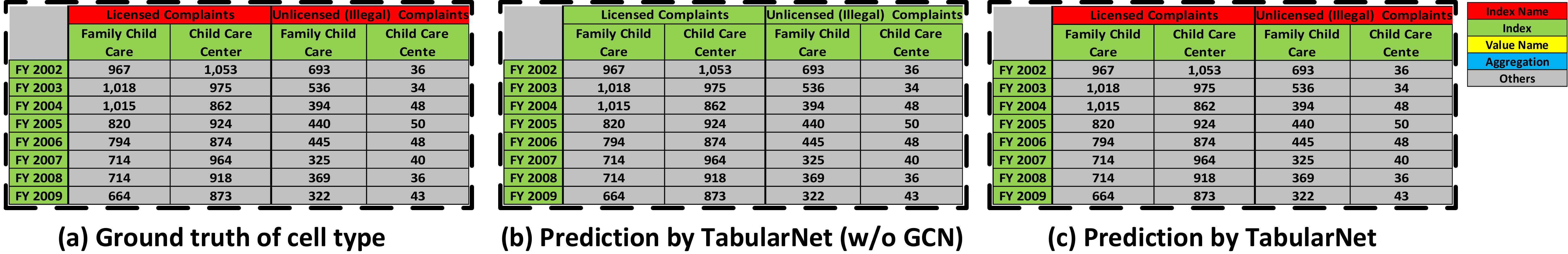}}
\caption{A case illustrating the effectiveness of the GCN module. (a) The ground truth of cells type in the Table, which is
indicated by different colors. (b) Cell types prediction by TabularNet which is emasculated by removing the GCN module. (c) Cell
types prediction by full suit TabularNet.} 
\label{fig:AblCase1}
\end{figure*}
\begin{figure*}
\centering
\scalebox{0.95}{
\includegraphics[width=1\textwidth]{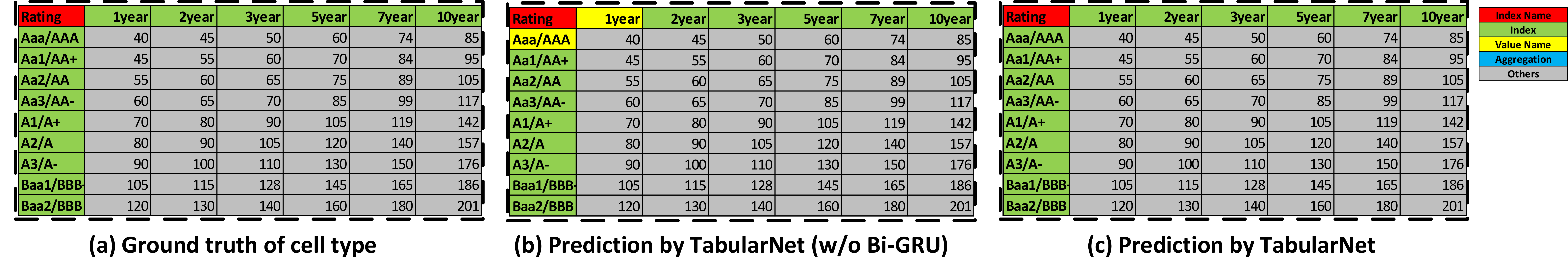}}
\caption{A case illustrating the effectiveness of the Bi-GRU module. (a) The ground truth of cells type in the Table, which is
indicated by different colors. (b) Cell types prediction by TabularNet which is emasculated by removing the Bi-GRU module. (c)
Cell types prediction by full suit TabularNet.} 
\label{fig:AblCase2}
\end{figure*}

\section{Evaluation}
\noindent\textbf{Q1: What is the effectiveness of our proposed model TabularNet?}
Table~\ref{cell-class-res} illustrates the evaluation results of various baselines on the table cell classification task. Macro-F1 score and F1-scores for four types of cell roles are reported
\footnote{Due to highly imbalanced class distribution, the F1-scores of the majority type \textit{Others} are similar among different models. Thus, we do not report the Micro-F1 score and the F1-score of this class.}. 
RNN-PE (Our features) is a variant of the original method with pretrain-embeddings replaced by features we proposed. From the results, we can observe that TabularNet outperforms all baselines. 
SVM and CART acquire competitive F1-scores on class \textit{Index} for the features we extracted are already sufficient in distinguishing \textit{Index} cells from \textit{Other} cells \cite{gol2019tabular}. 
However, the real difficulty lies in separating the remaining class \textit{Index Name}, \textit{Index}, and \textit{Value Name} from each other, which is extremely hard without incorporating tabular structure information (Fig. \ref{fig:Intro_task} (b)). 
On class \textit{Index Name} and \textit{Value Name}, RNN-PE, FCNN-MT and TAPAS significantly outperforms SVM and CART, because they all incorporate spatial information even though they are not so comprehensive compared with our model.

The structure information mining layer of TabularNet contains two vital operations: GCN and row/col-level Bi-GRU. As shown in Table~\ref{cell-class-res}, an ablation experiment is conducted to validate the necessity of each operation: TabularNet (w/o Bi-GRU) is a basic version that removes the Bi-GRU operation while TabularNet (w/o GCN) removes the GCN operation from the full suite model TabularNet. The superior of TabularNet (w/o Bi-GRU) over baselines verifies the rationality of capturing hierarchical and paratactic relationships of tabular data. The steady improvement in the performance indicates that all designs of TabularNet 
make their unique contributions.

\noindent\textbf{Q2: How generic is TabularNet for different tasks?}
We present the evaluation results of various models on the table region detection task in Table~\ref{region-class-res}. F1-scores of top and left header detection are reported. From Table~\ref{cell-class-res} we can observe similar comparison results, which demonstrates that TabularNet, as a common backbone, is effective for different tasks of table understanding, i.e., cell classification and region detection. 

The significant distinction between region detection and cell classification task is that they focus on different levels of the tabular structure. The former task relies more on the row/column-level structure, while the latter counts on the spatial structures of cells. That is the reason why TabularNet (w/o Bi-GRU) (can not capture much row/col-level information) fails to maintain advantages against baselines on this task. 

The validity of TabularNet promotes us to build a multi-task framework using TabularNet as the backbone. Intuitively, the training process of \textit{TabularNet (Multi-task)} utilizes data from both cell classification and region detection tasks. Therefore, it will enhance the generalization capability of our model. The intuition can be verified by the bolded figures in Table~\ref{cell-class-res} and Table~\ref{region-class-res}. In both tasks, \textit{TabularNet (Multi-task)} outperforms TabularNet that solely trained on a single task.

\noindent\textbf{Q3: Does the WordNet based graph construction algorithm outperforms the naive spatial-based construction method?}
Tabular data has two natural characteristics: 1) All the cells are arranged like a grid in two-dimensional space; 2) Data are ordered in two orthogonal directions, from top to bottom, and left to right side (TlBr). Therefore, we construct two kinds of basic graph structure directly:
\begin{itemize}
\item Grid: Cells are linked without direction to those that sharing one border. 
\item TlBr: Based on edges from Grid, only retain those edges with direction from top to bottom or left to the right.
\end{itemize}

Table \ref{tableedge} shows the influence of different graph construction methods on the cell classification task. The insufficiency of solely spatial relations and the effectiveness of WordNet edges can be verified from the superiority of heuristics-based results. 

\begin{figure*}
\centering
\scalebox{1.0}{
\includegraphics[width=1\textwidth]{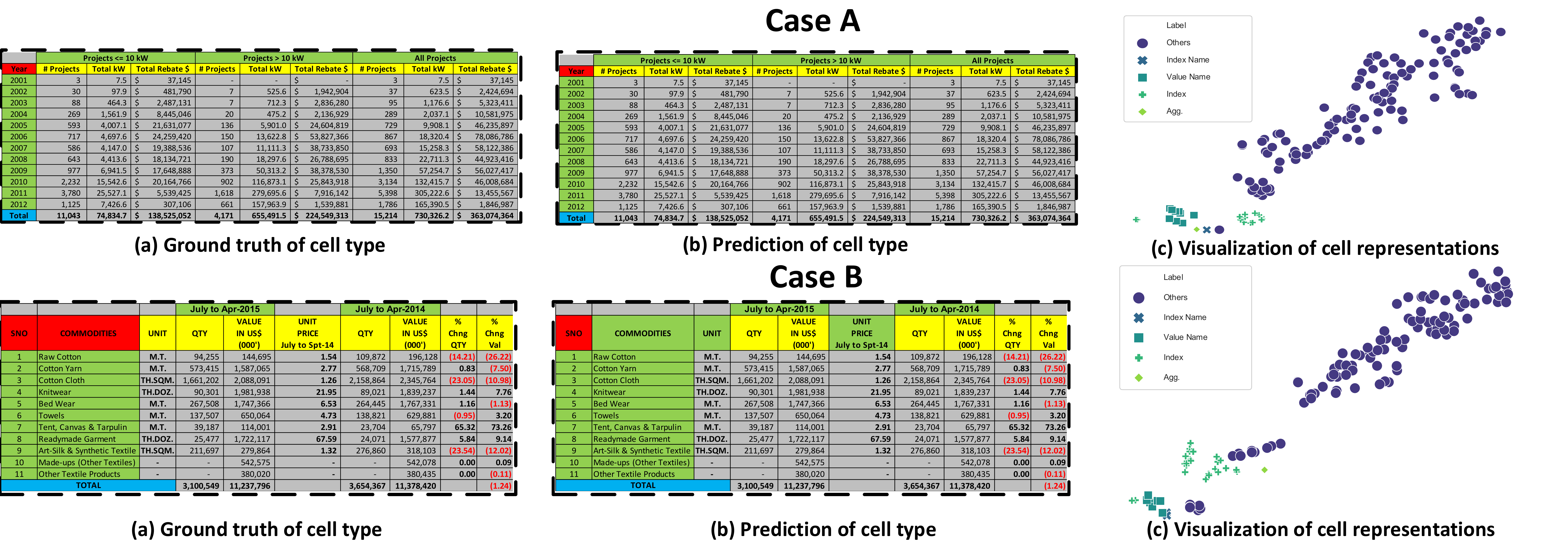}}
\caption{Two real cases for the cell classification task} 
\label{two:cases}
\end{figure*}

\noindent\textbf{Q4: How does TabularNet perform in real complicated cases?}
As shown in Fig. \ref{two:cases}, two cases are posted to illustrate the ability of cell type prediction of TabularNet. Some basic observations of those cases are:
\begin{itemize}
\item Both cases are cross tables, containing all kinds of cell types. They both have a hierarchical structure of 3 layers in their top region: One \textit{Index} cell leads several \textit{Value Name} cells, followed by the \textit{Other} cells. This kind of hierarchical structure is widespread in real tabular data.   
\item Two \textit{Index} cells lead each row of case B, while it is very delicate to regard the second cell as an Index even for humans.
\end{itemize}

Case B is more complex than A, and it has more ambiguous borders between the data region and the header region. What's worse, some columns are led by a single header cell. In each case, determining whether a cell is \textit{Index} or \textit{Value Name} must be based on a combination of its semantic information and the statistical information of the entire column. The prediction results of our method verify these statements:
\begin{itemize}
\item All the cells of case A are accurately predicted by our method, which shows that our method can handle a large portion of the real data similar to case A.
\item In case B, our method gives the right prediction to most of the cells in the second column. We attribute our success to two aspects: 1) the uniqueness (the prerequisite to be \textit{Index}) of cells in the whole column can be obtained in our TabularNet. 2) our model captures the global structure information that the bottom of the second column is merged into the aggregation cell \textbf{Total}, which is strong evidence for identifying the corresponding column as a header column. 
\item TabularNet fails to distinguish some cells between type \textit{Index} and type \textit{Value Name} in case B. As we discussed above, this is essentially a tricky task. It requires carefully balancing various information to make the final judgment, which is one direction that our model can be further improved. However, some proper post-processing rules can be used to fix this kind of misclassifications \cite{gol2019tabular}.
\end{itemize}

\begin{figure}
\centering
\scalebox{0.8}{
\includegraphics[width=0.46\textwidth]{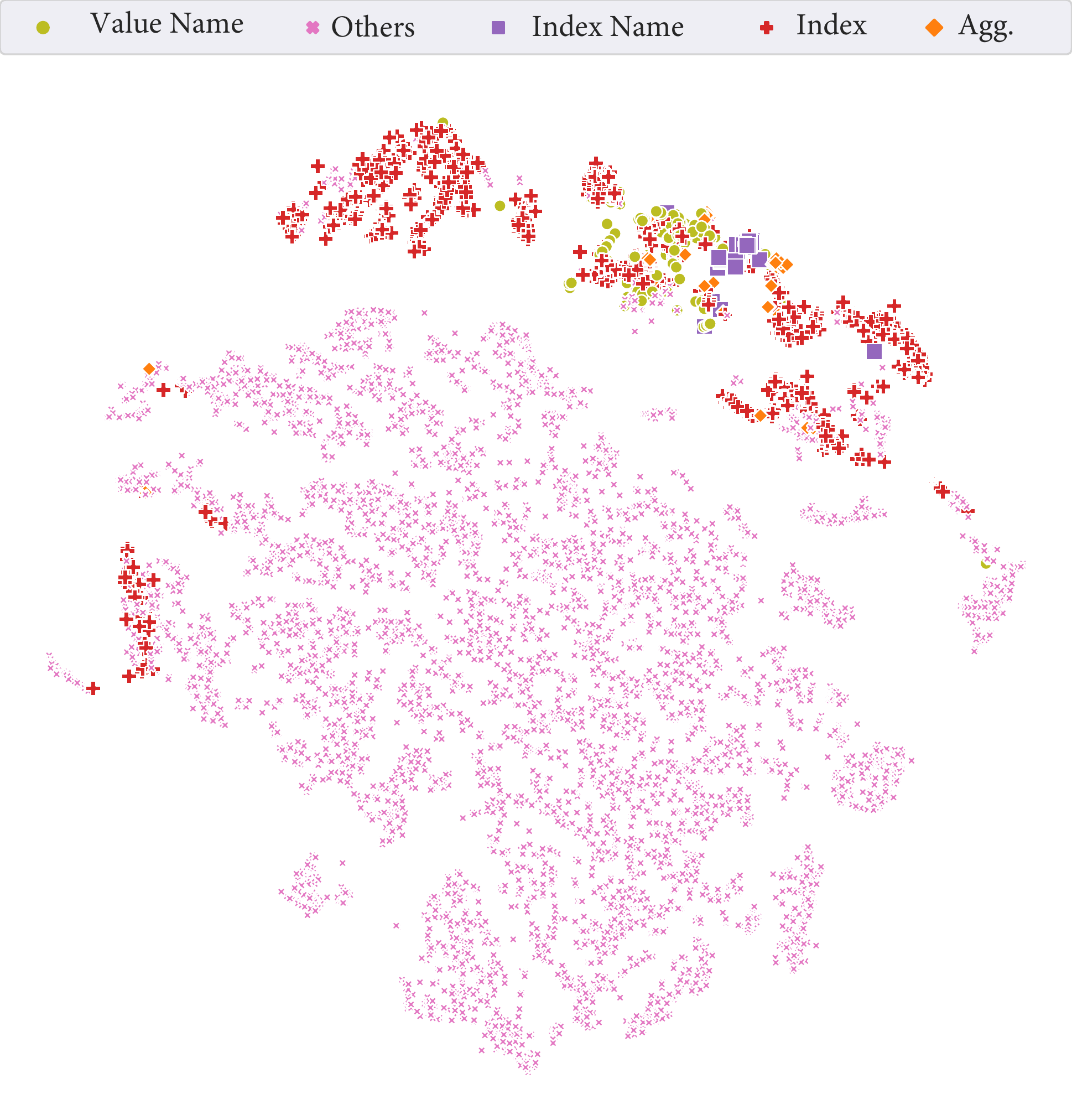}}
\caption{Embedding visualization of all cells in 50 randomly selected tables}
\label{emb:k}
\end{figure}

\noindent\textbf{Q5: Are the spatial and relational information necessary?}
In the Table \ref{region-class-res} and Table \ref{cell-class-res}, the ablation study shows that the GCN operation (to capture relational information) and Bi-GRU operation (to capture spatial information) are both necessary for improving the effectiveness of both tasks. To further understand how two kinds of information help the model, we present two real cases to show the capabilities of GCN and Bi-GRU modules, respectively.

It is shown that the usage of GCN can help us better distinguish Index Name and Index cells in Figure \ref{fig:AblCase1}. In this case, the first two rows have a hierarchical structure. The cells in the first row can be linked according to the Wordnet-based graph construction method, and then their embedding will be drawn close, which leads to a greater probability that the cells will be classified into the same class. Similarly, the cells in the second row are more likely classified into the same class. On the contrary, embeddings of cells in the first and the second row will be alienated since they do not have any links. Thus, using the GCN module can better distinguish cells in different hierarchical layers, and hence better classify Index and Index Name cells. In the second case (see Fig. \ref{fig:AblCase2}), the model without Bi-GRU misclassifies two cells as Value Name. Bi-GRU can capture the spatial relationship and make the classification results of the same row/col more consistent.

\noindent\textbf{Q6: Are the cell embeddings learned by TabularNet distinguishable?}
The effectiveness of the table's representation (i.e., embedding) determines the upper bound of the downstream task's performance. In TabularNet, we preserve the cell's vectorial representation before classifier as the cell's embedding. We utilize visualization to intuitively investigate the distinguishability of the learned embeddings. T-SNE \cite{maaten2008visualizing} is used for visualization exploring:
\begin{itemize}
\item In the right side of Fig. \ref{two:cases} shows the visualizations of case A and B, respectively. All cells in the header region are separated from data cells (type \textit{Other}). Cells of the same type are well clustered together with relatively little overlapping. 
\item To investigate the consistency of embedding across different tables, we randomly select 50 tables from the testing set and visualize all cells' embedding in one picture (see Fig. \ref{emb:k}).
Surprisingly, the distinction between the different cell types remains clear. Even the cells from different tables, of the same type, still gathered in the embedding space. 
\end{itemize}

\section{Conclusion}
    In this paper, we propose a novel neural architecture for tabular data understanding (TabularNet), modeling spatial information and diverse relational information simultaneously. Meanwhile, TabularNet can be viewed as a unified backbone for different understanding tasks at multiple scales, and hence can be utilized in a multi-task or a transfer learning scenario. We demonstrate the efficacy of TabularNet in cell-level, row/column-level, and multi-task classification in two real spreadsheet datasets. Experiments show that TabularNet significantly outperforms state-of-the-art baselines.

    In future work, we will conduct more experiments in more tasks on tabular data to further evaluate the representation ability of TabularNet. In addition, TabularNet can be combined with pretrained loss to better leverage the semantic and structural information from a large number of unlabeled tabular data.

\bibliographystyle{ACM-Reference-Format}
\bibliography{sample-base}

\clearpage
\appendix
\section{Feature Selection}
Cells in the spreadsheet have plentiful features, most of which are directly accessible. We can extract custom features from the inner content of cells. Based on features proposed by \cite{dong2019tablesense}, we build up our feature selection scheme, as shown in Table~\ref{feature:list}. We only elaborate on processing details of the following features as most features are plain:  
\begin{itemize}[leftmargin=10pt]
\item \textbf{Colors:} All colors are originally represented in the ARGB model.  To minimize the loss of information, we simply rescale color in all four channels, resulting in a four-dimensional vector.
\item \textbf{Coordinates:} The relative location of the cells on the table is critical. With the unknown size of the table, at least two anchor points are needed to indicate the position of a cell. Cells on the top left and bottom right are chosen as anchor points, which gives a four-dimensional position vector $[rt, ct, rb, cb]$ as shown in Table\ref{feature:list}.
\item \textbf{Decayed Row/Col Number:} One characteristic of tabular data is that the non-header rows and columns are usually permutation invariant, which indicates that the relative positions of cells in the data region are less critical. By an exponential decay function, we eliminate their position information in the data region while retaining it around the header region. Mathematically, $decayed_{row/col} = e^{-row/col}$, where $row$ and $col$ represents row and column number of a cell, respectively.
\item \textbf{Pretrained BERT:} To capture semantic information of a cell, we incorporate BERT to extract embeddings from their text \cite{dong2019semantic}. For cells without text, we simply use zero vectors as their semantic embeddings.
\end{itemize}

\vfill\eject 
\begin{table}[!ht]
\centering
\setlength{\abovecaptionskip}{5pt}%
\setlength{\belowcaptionskip}{5pt}%
\caption{Feature Selection Scheme}

\label{feature:list}
\resizebox{.9\linewidth}{!}{
\begin{tabular}{|l|l|l|}
\hline
\multicolumn{1}{|c|}{Feature Category} & \multicolumn{1}{c|}{Description} & \multicolumn{1}{c|}{Value} \\ \hline
\multicolumn{1}{|c|}{} & Length of text & int \\ \cline{2-3} 
\multicolumn{1}{|c|}{} & Is text empty? & bool \\ \cline{2-3} 
\multicolumn{1}{|c|}{Text} & Ratio of digits & float \\ \cline{2-3} 
\multicolumn{1}{|c|}{} & If "\%" exits & bool \\ \cline{2-3} 
\multicolumn{1}{|c|}{} & If "." exits & bool \\ \hline
\multicolumn{1}{|c|}{} & Is number? & bool \\ \cline{2-3} 
\multicolumn{1}{|c|}{} & Is datatime? & bool \\ \cline{2-3} 
\multicolumn{1}{|c|}{Text Format} & Is Percentage? & bool \\ \cline{2-3} 
\multicolumn{1}{|c|}{} & Is Currency? & bool \\ \cline{2-3} 
\multicolumn{1}{|c|}{} & Is Text? & bool \\ \cline{2-3} 
\multicolumn{1}{|c|}{} & Others & bool \\ \hline
& FillColor & {[}R,G,B,A{]}/255 \\ \cline{2-3} 
& TopBorderStyleType not None & bool \\ \cline{2-3} 
& BottomBorderStyleType not None & bool \\ \cline{2-3} 
& LeftBorderStyleType not None & bool \\ \cline{2-3} 
\multicolumn{1}{|c|}{Cell Format} & RightBorderStyleType not None & bool \\ \cline{2-3} 
& TopBorderColor & {[}R,G,B,A{]}/255 \\ \cline{2-3} 
& BottomBorderColor & {[}R,G,B,A{]}/255 \\ \cline{2-3} 
& LeftBorderColor & {[}R,G,B,A{]}/255 \\ \cline{2-3} 
& RightBorderColor & {[}R,G,B,A{]}/255 \\ \hline
& FontColor & {[}R,G,B,A{]}/255 \\ \cline{2-3} 
\multicolumn{1}{|c|}{Font format} & FontBold & bool \\ \cline{2-3} 
& FontSize & float \\ \cline{2-3} 
& FontUnderlineType not None & bool \\ \hline
& Height/width (rescaled) & float \\ \cline{2-3} 
& HasFormula & bool \\ \cline{2-3} 
& IndentLevel & float \\ \cline{2-3} 
\multicolumn{1}{|c|}{Others} & Coordinates & {[}rt, ct, rb, cb{]} \\ \cline{2-3} 
& Decayed row number & float \\ \cline{2-3} 
& Decayed column number & float \\ \hline
Text Embedding & Pretrained BERT & Embedding of sise 768 \\ \hline
\end{tabular}
}
\end{table}

\section{Implementation Details}
We implement our model utilizing Pytorch\footnote{https://github.com/pytorch/pytorch} and Pytorch-Geometric \cite{Fey/Lenssen/2019}. We use Glorot~\cite{glorot2010understanding} method (Xavier Normal) to initialize all parameters in our model. The Bi-GRU modules consist of 3 recurrent layers. We stack 2 layers of GIN for relational information encoding.

\end{document}